\documentclass[conference]{IEEEtran}
\usepackage{booktabs}
\usepackage{cite}
\usepackage{amsmath,amssymb,amsfonts}
\usepackage{graphicx}
\usepackage{textcomp}
\usepackage{xcolor}
\usepackage{url}
\usepackage{algorithm}
\usepackage{algpseudocode}

\DeclareMathOperator*{\argmax}{\arg\,max}

\begin{document}

\title{Robot See, Robot Do: Imitation Reward for Noisy Financial Environments}

\author{\IEEEauthorblockN{Sven Goluža\IEEEauthorrefmark{1}, Tomislav Kovačević\IEEEauthorrefmark{2}, Stjepan Begušić\IEEEauthorrefmark{3}, Zvonko Kostanjčar\IEEEauthorrefmark{4}}
\IEEEauthorblockA{\textit{University of Zagreb Faculty of Electrical Engineering and Computing} \\
\textit{Laboratory for Financial and Risk Analytics}\\
Unska 3, 10000 Zagreb, Croatia \\
\IEEEauthorrefmark{1}sven.goluza@fer.unizg.hr,
\IEEEauthorrefmark{2}tomislav.kovacevic@fer.unizg.hr,
\IEEEauthorrefmark{3}stjepan.begusic@fer.unizg.hr, and
\IEEEauthorrefmark{4}zvonko.kostanjcar@fer.unizg.hr}}

\maketitle

\begin{abstract}
The sequential nature of decision-making in financial asset trading aligns naturally with the reinforcement learning (RL) framework, making RL a common approach in this domain. However, the low signal-to-noise ratio in financial markets results in noisy estimates of environment components, including the reward function, which hinders effective policy learning by RL agents. Given the critical importance of reward function design in RL problems, this paper introduces a novel and more robust reward function by leveraging imitation learning, where a trend labeling algorithm acts as an expert. We integrate imitation (expert's) feedback with reinforcement (agent's) feedback in a model-free RL algorithm, effectively embedding the imitation learning problem within the RL paradigm to handle the stochasticity of reward signals.  Empirical results demonstrate that this novel approach improves financial performance metrics compared to traditional benchmarks and RL agents trained solely using reinforcement feedback.
\end{abstract}


\section{Introduction}
Financial asset trading is well-suited to the reinforcement learning (RL) framework, where an agent learns through trial and error by interacting with its environment and receiving feedback on its actions \cite{Sutton2018}. As financial markets are notorious for their non-stationarity and low signal-to-noise ratio \cite{Black1986}, they present a unique challenge for applying RL to learn trading strategies. A crucial aspect of the RL setup is designing the reward function, which provides the agent with necessary feedback on its performance \cite{Ng2000}. Consequently, managing noisy samples from the underlying reward function of the financial environment presents a significant problem that is often overlooked \cite{Romoff2018}.

In general, the reward function in RL represents the utility function that quantifies the agent’s preferences for different outcomes. Designing the reward function is challenging because it needs to be efficient and easy to optimize, as well as convey the desired task \cite{Ding2019}. In financial applications, the reward function is usually designed in terms of the agent's profit and the risk taken to realize that profit. However, the high degree of stochasticity in noisy market environments often leads to a high degree of stochasticity in the training process and policy learning. Most research has focused on dealing with this noise by improving state and policy representations \cite{Deng2016, Li2019, Liu2020}. However, less attention has been paid to addressing noise specifically within the reward signal itself \cite{Everitt2017}, despite the fact that profit-based reward signals are noisy as it is often difficult to distinguish whether the gains or losses are genuinely driven by underlying price movements or are just random fluctuations around the actual price. In \cite{Sun2022}, the authors introduce a reward function that combines the agent's profit with a hindsight bonus to incentivize long-term trading decisions. This approach mitigates the noise associated with short-term price fluctuations in the agent's profit and demonstrates the agent's robust performance. In \cite{Liu2020}, an imitation learning technique is employed where the expert takes a long position at the lowest price and a short position at the highest price within a given day. These expert actions are incorporated into the agent's policy learning to reduce the inefficient exploration phase. By combining the agent's reward (profit) with the expert's actions, the authors reduce the exploration phase and show a more robust performance.

Previous research on stochasticity and corruption of the reward signal has shown that without simplifying assumptions, RL agents cannot be expected to avoid this problem \cite{Wang2020}. Therefore, providing agents with different data sources and feedback is often the safest option \cite{Everitt2017}. We build on these findings by using a specific trend labeling algorithm that introduces our agent with expert feedback and provides a more robust source of information compared to the previously mentioned methods. There are a variety of methods that deal with the stochasticity of reward signals, including those that use human feedback \cite{Christiano2017}, reward signal estimators \cite{Romoff2018, Hadfield-Menell2017}, distributional reward estimators \cite{Bellemare2017}, and other data-driven methods such as imitation learning (IL) \cite{Zheng2021}, where the agent learns to imitate the expert's behavior based on demonstrations. Our work is inspired by methods that reduce IL to an RL problem without explicitly learning the reward function, but inferring the reward signal from the expert's demonstrations \cite{Reddy2019, Ciosek2022, Luo2023}. This simple approach rewards the agent for matching the expert's action in a given state and punishes it otherwise. This encourages the agent to align itself with the expert's demonstrations over a long horizon \cite{Reddy2019}.

The key contribution of this paper is the introduction of a novel reward function that leverages IL and uses a trend labeling algorithm as an expert. This means that the agent learns not only from its own experience (reinforcement feedback) but also by mimicking the expert's actions (imitation feedback). By combining both sources of information in a model-free RL framework, the agent gains a more robust understanding of the market. We empirically validate our approach on intraday futures data and show significant improvements in risk-adjusted performance metrics compared to benchmarks consisting of traditional methods and RL agents trained solely using reinforcement feedback.

\section{Background}
In this section, we formalize the RL and IL frameworks in terms of a Markov decision process (MDP). We also introduce the trend labeling algorithm used as the expert. 
\subsection{Reinforcement learning}

The environment within which the RL agent interacts is described formally using MDP. The MDP is a tuple $\langle \mathcal{S}, \mathcal{A}, \mathcal{P}, \gamma, \mathcal{R}, p_0 \rangle$:
\begin{itemize}
        \item $\mathcal{S}$ is a set of states,
        \item $\mathcal{A}$ is a set of actions,
        \item $\mathcal{P}$ is a matrix of transition probabilities $p(s_{t+1}|s_t,a_t)$,
        \item $\gamma \in[0,1]$ is a discount factor,
        \item $\mathcal{R} : \mathcal{S} \times \mathcal{A} \times \mathcal{S} \mapsto \mathbb{R}$ is a reward function,
        \item $p_0$ is a distribution of initial states.
\end{itemize}
A policy $\pi$ is a mapping from state to action (or a distribution over actions). At each time step $t$, the agent performs an action $a_t \sim \pi(a_t|s_t)$ based on the current state $s_t$ of the environment. In return, the environment provides the agent feedback in the form of the reward signal $r_{t+1}$ and the agent switches to the next state $s_{t+1}$, which is determined by the transition probability matrix $\mathcal{P}$. An episode is a sequence consisting of states and actions $\tau = (s_0,a_0, s_1,a_1,...)$, and is also referred to as a trajectory. The length of this sequence is either fixed as in finite-horizon problems or arbitrary as in infinite-horizon problems. In this research, we assume a finite-horizon setup, where the agent makes decisions over one trading day, so we do not apply discounting of future rewards since the discount factor $\gamma$ is fixed to 1. The agent's goal is to find $\pi$ that maximizes the sum of future rewards $\sum_{t=i}^{T} r_i$ over the episode of length $T$:
\begin{equation}
    \pi^* = \argmax_{\pi_\theta} \mathbb{E}_{\tau \sim p_{\theta}(\tau)}[R(\tau)].
\label{eq:pi_opt}
\end{equation}
The agent's policy $\pi_\theta$ is parameterized by $\theta$, while $p_{\theta}(\tau)$ is a parameterized probability of a trajectory induced by the policy $\pi_{\theta}$. The expected value of the cumulative reward over the entire trajectory under the same policy is represented by $\mathbb{E}_{\tau \sim p_{\theta}(\tau)}[R(\tau)]$. RL algorithms solve problems formulated in terms of MDP and return the policy $\pi^*$.

The action-value function, also called Q-function, of a policy $\pi$ is defined as $Q^{\pi} (s_t, a_t) = \mathbb{E}_\pi[\sum_{t=i}^{T} r_i | s_t, a_t]$. The state-value function, also known as the V-function, is defined as $V^{\pi} (s_t) = \mathbb{E}_\pi[\sum_{t=i}^{T} r_i | s_t]$. These functions represent the expected cumulative reward, starting from a given state (V-function) or state-action pair (Q-function) and performing actions based on the policy $\pi$ for the rest of the trajectory. The advantage function, defined as $A^{\pi} (s_t, a_t) = Q^{\pi} (s_t, a_t) - V^{\pi} (s_t)$, indicates whether the action $a_t$ is better or worse than the average action performed by the policy $\pi$ in $s_t$. In deep reinforcement learning (DRL), policy, value, and advantage functions are all trained using reward signals and are typically represented as neural networks.

In this study, we conceptualize the trading problem for a single asset as an MDP, outlining each component (state, action, and reward function) in the following section.

\subsection{Imitation learning}
Rather than relying on trial-and-error learning and designing the reward function, IL provides a method for training a policy using a static dataset \cite{Goluza2023}. This dataset includes expert demonstrations represented as a set of expert trajectories $\{s_i,a_i\}_{i=0}^N$, and does not include any information about the reward signal. This approach is particularly useful in domains where it is more practical and easier to train the agent using demonstrations rather than defining a reward function and learning through interaction with the environment.

Learning the policy through supervised learning, where states are treated as features and actions as labels -- also known as behavioral cloning (BC) -- represents a naive approach. The major weakness of this method is the distributional shift caused by the agent's inability to generalize effectively. When the agent deviates from the demonstrated states, it starts to encounter unfamiliar states, leading to an accumulation of errors as these states were not encountered during training \cite{Ross2011}. This weakness is addressed by interactively querying the expert to label the states visited by the agent during deployment. The policy is then retrained using this new data, improving its performance in previously unseen states. There are different ways of utilizing the querying of the expert \cite{Kelly2019, Hoque2023}, but the idea of the expert guiding the agent remains the same.

Instead of learning the policy through supervised learning, methods based on inverse reinforcement learning (IRL) derive the reward function from expert demonstrations and then use the RL framework to maximize these rewards \cite{Ng2000}. However, since IRL comes with its own challenges and limitations, especially in noisy environments, the most scalable and versatile IRL methods are adversarial imitation learning methods. These methods use an adversarial training setup \cite{Ho2016, Fu2017}, where a discriminator is trained to minimize the divergence between the distribution of the expert demonstrations and the trajectories generated by the agent. The key idea in these methods is to incentivize the agent to return to the demonstrated states when it encounters new out-of-distribution states \cite{Reddy2019}. Due to the complexity of approximation techniques and the difficulty of training adversarial models, where training is often unstable, recent research \cite{Reddy2019, Wang2019, Ciosek2022, Luo2023} has introduced approaches that reduce IL to an RL problem by defining the reward function through expert demonstrations. These approaches offer simpler yet effective alternatives to complex adversarial methods.

Motivated by methods that reduce IL to an RL problem, we incorporate expert demonstrations directly into the reward function. The expert is represented by a trend labeling algorithm with access to future data. Using this approach, we demonstrate that the agent can learn more effective policies by integrating an additional, noise-free data source.

\subsection{Trend labels as expert demonstrations}
To enable the agent to learn expert behavior without introducing human bias, we generate the reward signal using an automated trend labeling algorithm. Instead of relying on real expert demonstrations and sentiment, which could affect the robustness of trading strategies, we employ an oracle labeling algorithm \cite{kovacevic2023}. This algorithm identifies optimal positions within the price time series that yield theoretically maximal cumulative returns, accounting for commission costs associated with initiating new positions. Providing significant and stable trend labels, oracle algorithm enhances the agent's ability to learn effective trading strategies across diverse market environments.

Consider a price time series $ \boldsymbol{p} = \{p_{t}\}_{t=1}^{T}, p_t \in \mathbb{R}^{+}$. Using a trend labeling algorithm $\boldsymbol{g}$, we can generate a label series $y_t \in \{0,1\}$ as $\boldsymbol{y} = \boldsymbol{g}(\boldsymbol{p}, \boldsymbol{\vartheta)}$, where $\boldsymbol{\vartheta}$ represents the algorithm parameters. The labels themselves have a clear trading interpretation -- a transition from $0 \rightarrow 1$ indicates an uptrend, suggesting the initiation of a long position. Conversely, a transition from $1 \rightarrow 0$ indicates a downtrend and closing a long position. The return of the $i$-th position can be calculated as:
\begin{equation}
    r_i = \frac{p_{t+\Delta t} - p_t \cdot (1 + \Theta)}{p_t \cdot (1 + \Theta)},
\label{eq:pos_return}
\end{equation}
where $\Delta t$ represents the holding period of a position (from $0 \rightarrow 1$ until $1 \rightarrow 0$), and $\Theta$ represents the  commission cost for opening a long position in the asset. Examples of two positions yielding $r_1$ and $r_2$ respectively can be found in Fig.~\ref{fig:positions}. Note that
$r_i$ in this subsection refers to the return of a single position for the purpose of introducing the trend labeling algorithm, while $r$ in the following sections refers to the reward signal in the context of RL. The cumulative return $r_{cum}=c(\boldsymbol{p}, \boldsymbol{y},\Theta)$ from $S$ long positions is calculated as:
\begin{figure}[b]
    \centering
    \includegraphics[width=0.49\textwidth]{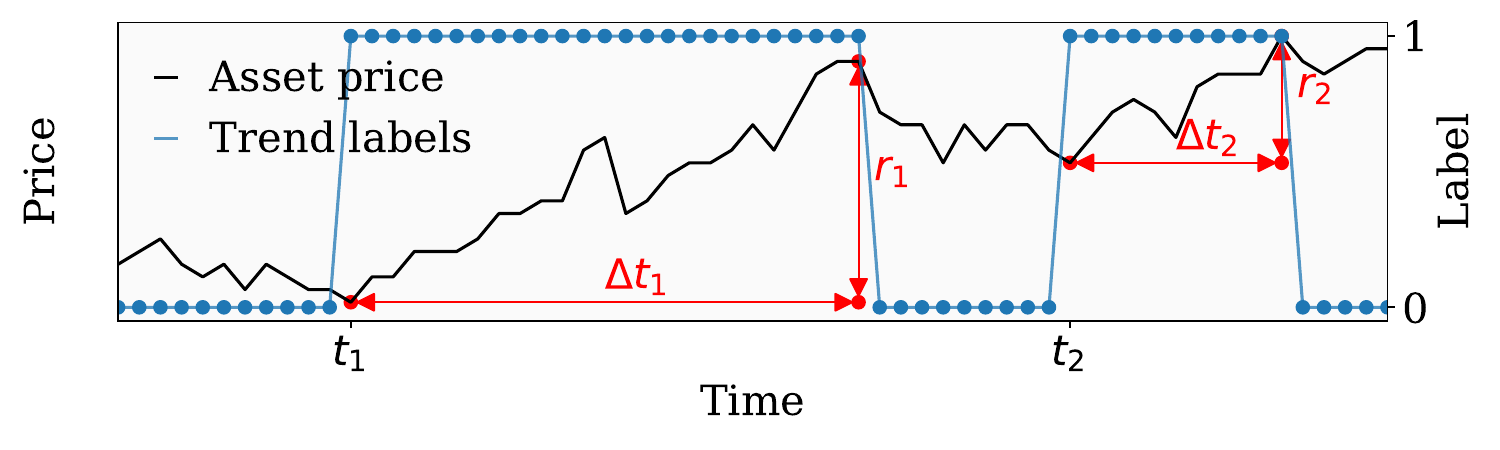}
    \caption{Two positions with $\Delta t_1$ and $\Delta t_2$ holding periods yielding $r_1$ and $r_2$ returns, respectively, as a result of two labeled uptrends.}
    \label{fig:positions}
\end{figure}
\begin{equation}
    r_{cum} = \prod_{i=1}^{S} (1 + r_i) - 1,
\label{eq:cum_ret}
\end{equation}

Oracle labels are obtained by solving the following optimization problem:
\begin{align}
    &\boldsymbol{y}^* = \arg\max_{\boldsymbol{y}} c(\boldsymbol{p}, \boldsymbol{y}, \vartheta) \\ \nonumber
    &\text{subject to:} \\ \nonumber
    &\quad \vartheta = \Theta, \\ \nonumber
    &\quad y_T = i, \quad i \in \{0, 1\},
\label{eq:oracle_optimisation}
\end{align}
where $y_T$ represents the final label and $\vartheta$ represents the commission costs. This optimization can be solved with a dynamic programming approach as described in Algorithm \ref{alg:oracle_labeling_alg}. The state matrix represents the maximum cumulative return in time step $t$ while being in the state $i\in\{0,1\}$. The transition cost matrix corresponds to the change in the cumulative return due to the transition from state $i$ at time step $t-1$ to state $j$ at time step $t$.

\begin{algorithm}
\caption{Oracle labeling algorithm.}
\label{alg:oracle_labeling_alg}
\begin{algorithmic}[1]
\State \textbf{Input}: $\boldsymbol{p}$, $\vartheta$, $y_{T}$
\State \textbf{Output}: $\boldsymbol{y}$
\State $\boldsymbol{S}\gets$ \texttt{init\_state\_matrix}$(0, y_{T}, T)$
\State $\boldsymbol{P}\gets$ \texttt{init\_transition\_cost\_matrix}$(\boldsymbol{p}, \vartheta, T)$
\For{$t \gets 2$ \textbf{to} $T$}
\State $S_{t}^{0} \gets \max \left( S_{t-1}^{0} + P_{t-1}^{0, 0}, S_{t-1}^{1} + P_{t-1}^{1, 0} \right)$
\State $S_{t}^{1} \gets \max \left( S_{t-1}^{0} + P_{t-1}^{0, 1}, S_{t-1}^{1} + P_{t-1}^{1, 1} \right)$
\EndFor
\State $\boldsymbol{y}\gets$ \texttt{init\_label\_array}$(y_{T}, T)$
\State $\kappa\gets y_{T}$
\For{$t\gets T-1$ \textbf{to} $1$ \textbf{by} $-1$}
\State $idx\gets \arg\max_{i \in \{0,1\}}\left(S_{t}^{i} + P_{t}^{i, \kappa}\right)$
\State $y_{t} \gets idx$
\State $\kappa\gets idx$
\EndFor
\State \textbf{return} $\boldsymbol{y}$
\end{algorithmic}
\end{algorithm}

\begin{figure}[h!]
    \centering
    \includegraphics[width=0.49\textwidth]{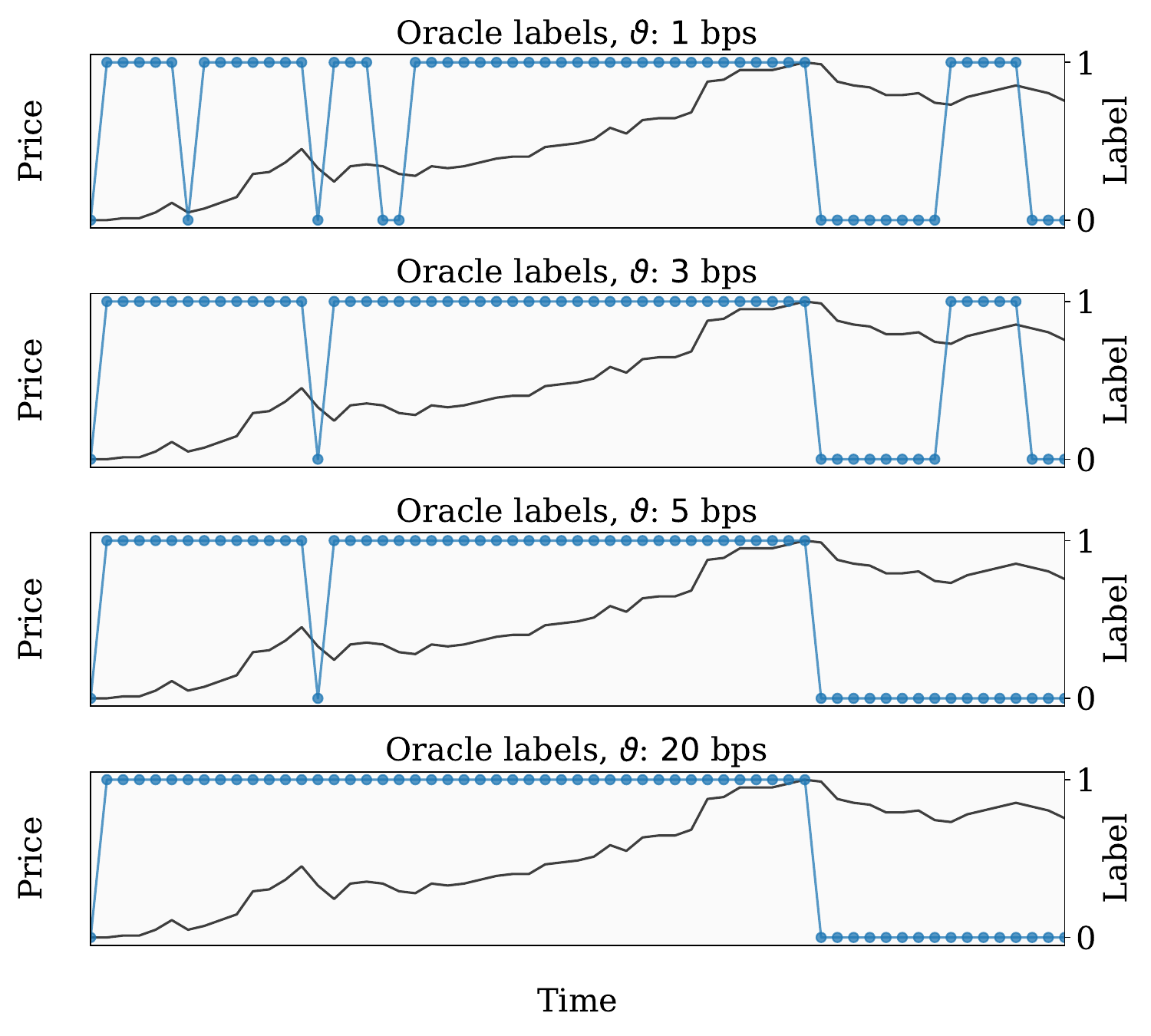}
    \caption{Four different sets of labels obtained using the oracle labeling algorithm, each considering a different level of commission costs, expressed in basis points (bps).}
    \label{fig:oracle_labels}
\end{figure}

An example of labels obtained with this algorithm for different commission costs $\vartheta$, expressed in basis points (bps), is shown in Fig.~\ref{fig:oracle_labels}. It is evident that a small $\vartheta$ leads to a large number of positions with a short holding period, as commission costs have a minimal impact on cumulative return. On the other hand, as $\vartheta$ increases, the number of profitable positions decreases, and positions are fewer in number and held for longer periods. If the asset return does not surpass the predefined commission costs, no positions are taken.

\section{Methodology}
In this section, we present the individual components of the MDP along with the model-free learning algorithm. The MDP was developed for an episodic intraday trading task, where the agent makes minute-level decisions throughout the trading day and liquidates positions by the end of the day to mitigate overnight risk.

\subsection{State space}
The state space is represented by price-based features and features derived from the agent's positional context \cite{Goluza2024}. The price-based features include six technical indicators: \emph{Williams \%R}, \emph{relative strength index}, \emph{commodity channel index}, \emph{ultimate oscillator}, and \emph{average directional index}. These indicators aim to capture distinct characteristics of the price time series \cite{Murphy1999}. Additionally, the state space includes the agent's position and the remaining time until the trading day concludes, forming a state vector $s_t \in \mathbb{R}^8$. Each continuous component within the state vector is normalized using min-max normalization, ensuring that all features lie within the interval $[-1,1]$.

\subsection{Action space}
The discrete action space consists of two possible actions, $a_t \in \{0,1\}$, which represent the position the agent can take, rather than the decision to buy or sell \cite{Zhang2020}. Since we focus on integrating oracle labels into the reward function, which are easier to interpret in a scenario with binarized labels, we analyze long-only strategies where the agent can only take long positions.

\subsection{Reward function}
The primary focus of this study revolves around introducing a novel reward function. Before formulating the reward function, we define the relevant terminology. The price vector, consisting of open, high, low, and close prices observed within a one-minute interval, is denoted as $\mathbf{p}_t = [p_t^\text{O}, p_t^\text{H}, p_t^\text{L}, p_t^\text{C}]$. Each one-minute interval is indexed by the discrete time step $t$, which represents the conclusion of the interval. At each time step $t$, the agent has access to all price information, extending up to $\mathbf{p}_{t}$.

To incorporate trading costs into our simulation, we consider commission costs. The commission, denoted as $c$, is calculated as a percentage (denoted as $\phi$) of the traded price. We assume that trades are executed at the open price of the subsequent minute. Given that our trading strategy involves unit size positions exclusively, any market impact is considered negligible and thus does not affect price movement in our model.

The proposed reward function consists of two main components: reinforcement feedback and imitation feedback. The reinforcement feedback $r^{\text{RF}}$ is defined as the additive profit at each time step:
\begin{align}
    &r^{RF}_{t+1} = a_t \cdot (p_{t+1}^\text{C} - p^{\text{exec}}) - c(a_t,a_{t-1};\phi)\\
    & c(a_t,a_{t-1};\phi) = \phi \cdot p_{t+1}^\text{O} \cdot |a_t - a_{t-1}|\\
    &p^{\text{exec}} = \begin{cases}
    p_{t+1}^{\text{O}},& \text{if } |a_{t} - a_{t-1}|\neq 0\\
    p_{t}^{\text{C}},& \text{otherwise.}
\end{cases}
\end{align}
The commission cost $c$, which is parameterized by $\phi$, depends on both the current and previous actions taken. The execution price, $p^{\text{exec}}$, represents the open price if the agent makes trades in that time step, or the previous close price if the agent maintains its position. 

The imitation feedback $r^{\text{IF}}$ is defined as the additive profit generated by the oracle labeling algorithm, which utilizes the oracle labels $y_t$ at each time step: 
\begin{align}
    &r^{IF}_{t+1} = y_t \cdot (p_{t+1}^\text{C} - p^{\text{exec}})\\
    &p^{\text{exec}} = \begin{cases}
    p_{t+1}^{\text{O}},& \text{if } |y_{t} - y_{t-1}|\neq 0\\
    p_{t}^{\text{C}},& \text{otherwise.}
\end{cases}
\end{align}

Finally, combining the reinforcement feedback $r^{RF}$ and imitation feedback $r^{IF}$, the proposed reward function $r^{RIF}$ is constructed as:
\begin{equation}
    r_{t+1}^{RIF} = r^{RF}_{t+1} - r^{IF}_{t+1}.
\end{equation}
Subtracting the imitation feedback from the reinforcement feedback can be interpreted as incorporating a baseline into the reward signal, which leads to more stable reward signals. This helps in reducing the variance of the noisy gradients and makes the training process more robust and effective.

The proposed reward function offers distinct advantages over traditional noisy reward functions using solely profit-based feedback. By incorporating imitation feedback derived from labels, our approach alleviates the noise inherent in reward signals influenced by price fluctuations. This ensures that the agent not only seeks immediate profit opportunities but also learns from stable and consistent strategies, obtained from the labeling algorithm. Consequently, the reward function guides the agent towards actions that align more closely with identified market trends, enhancing learning efficiency and overall performance in dynamic environments. By emphasizing both profitability and alignment with the oracle labels, our approach addresses the challenge of distinguishing actionable market signals from noise.

In practical terms, this reward function facilitates three outcomes. Firstly, it enables the agent to learn from oracle labels by providing a near-zero reward when the agent matches the actions implied by the labels, akin to supervised learning but adapted to consider the sequential nature of state and action interactions. Secondly, it encourages the agent to develop a better policy than the "expert" by rewarding the agent for exploiting small price fluctuations when the oracle label is out of position. Lastly, it ensures the agent can recover from losses by penalizing it when it deviates from the oracle labels, thereby guiding the agent back to the "expert" strategy. 

\subsection{Learning algorithm}

The proximal policy optimization (PPO) algorithm \cite{Schulman2017} is employed to train the agent's policy. This model-free DRL algorithm works directly in the policy space and eliminates the need to learn the transition dynamics of the trading environment. The core idea involves optimizing a surrogate objective function while ensuring the policy remains close to its previous iteration. To avoid significant policy changes in successive time steps, the surrogate objective is clipped:
\begin{align}
    L_t(\theta) &= \hat{\mathbb{E}}_t[\text{min}(\rho_t(\theta) \hat{A}_t, \text{clip}(\rho_t(\theta), 1 - \epsilon, 1 + \epsilon)\hat{A}_t)] \label{eq:clip_loss}, \\
    &\text{where:} \nonumber\\
    &\quad \rho_t(\theta) = \frac{\pi_{\theta_i(a_t | s_t)}}{\pi_{\theta_{i-1}(a_t | s_t)}}, \\
    &\quad \text{clip}(x, \text{min}, \text{max}) =
\begin{cases} 
\text{min} & \text{if } x < \text{min} \\
\text{max} & \text{if } x > \text{max} \\
x & \text{otherwise}
\end{cases}.
\end{align}
Empirical expectation over a finite batch of samples is signified by $\hat{\mathbb{E}}_t$.  The state-value function is designed as a neural network that shares all layers with the policy $\pi_\theta$, except the final layer, which outputs the state value rather than action probabilities. This state-value function is used in calculating the advantage function $\hat{A}_t$. 

The clipped objective ensures stability and efficient reuse of experience samples, facilitating multiple gradient steps on the same mini-batch, while conservative policy updates ensure the learning process is robust. These qualities, along with its simplicity and sample efficiency, make PPO a crucial component of our RL framework.

\section{Experimental setup}
In this section we outline the data used in our experiments, describe the trading environment and the training processes in detail. The experimental setup is intentionally kept simple and straightforward to emphasize the fundamental ideas and effectiveness of the proposed approach. Extensions and additional specifications, such as expanding the action space are beyond the scope of this study.

\subsection{Data and preprocessing}
We use a futures dataset that includes three assets historically known for their liquidity and sufficient order flow: GC (Gold), CL (Crude Oil), and ES (S\&P 500). The dataset consists of OHLCV data sampled at one-minute intervals during the most liquid trading hours, from Monday to Friday, 09:30 to 17:00 ET, based on information from the exchange platform.

To train and test the agent, we adopt a rolling window approach over the dataset. Each rolling window includes a training period of one year, followed by a validation period of three months and a testing period of three months. During the validation period, hyperparameters related to the reward function are tuned, and early stopping techniques are applied to halt agent training. Other hyperparameters remain fixed for faster experimentation.

The testing phase covers 18 months of minute-level data (October 2022 to March 2024), ensuring that our experiments evaluate performance across diverse market conditions. During this period, each asset exhibited distinct market behaviors under an intraday trading strategy of buying at the beginning and selling at the end of the trading day. Specifically, ES (S\&P 500) was in a bullish state, CL (Crude Oil) was in a bearish state, and GC (Gold) demonstrated mean-reverting behavior. By including assets from different classes—metal, energy, and equity index—each driven by unique factors and exhibiting varying market states, our experiments allow for our agent to be tested comprehensively.

\subsection{Environment and training details}
Since we model the trading problem as a Markov Decision Process (MDP) for an episodic task, each episode corresponds to one trading day. As previously mentioned, we consider only the liquid trading hours ($09$:$30$ -- $17$:$00$ ET). To construct the features for the state space, the agent requires a 61-minute lookback window, enabling trading to begin at $10$:$32$. The agent closes any open positions at the end of the trading day, exiting at $16$:$58$ to mitigate the potential impact of significant volatility during the final minute of trading hours.

Before training, we ensure that our state space features are properly scaled using min-max normalization, which scales each numerical feature to the interval $[-1,1]$. Our neural networks for the policy and value functions are designed with two hidden layers: one with 64 units and another with 32 units. We use the $\tanh$ activation function in these shallow neural networks, set a learning rate of $\alpha = 0.0001$, and employ the Adam optimizer for parameter optimization. During each iteration of the algorithm, we perform $10$ epochs of gradient updates using an experience buffer with $1024$ samples. Within each epoch, we split the data into mini-batches of size $64$ for efficient computation. 

Hyperparameter optimization is conducted using a validation set that spans three months following the training period. We specifically optimize the commission fee $\vartheta$ in the imitation feedback and the commission fee $\phi$ in the reinforcement feedback. We perform a grid search, using a search space consisting of the values $\{0.5, 1, 2, 3, 4, 5, 10, 20\}$ bps for each hyperparameter. The values that yield the best performance on the validation set, in terms of cumulative return, are selected. Evaluation statistics on the validation and test sets are computed using the reinforcement feedback of the agent with the commission fee of $1$ bps. All other hyperparameters are fixed and listed above.

After completing the training for each day within the training period, we evaluate the performance of the agent on the validation set. If the performance does not improve in three consecutive epochs, we stop the training of the agent early using an early stopping mechanism.

\section{Results and discussion}

\renewcommand{\arraystretch}{1.2}
\setlength\tabcolsep{10pt}
\begin{table*}[t]
\centering
\begin{tabular}{r|rrrrr}
\toprule
& \multicolumn{5}{c}{GC} \\
& num. trades & winrate[\%] & $\hat{\mu}_{r>0}$[\%] & $\hat{\mu}_{r<0}$[\%] & avg. holding period [min.] \\ \hline
\textit{RIF}          & 231 & 49.351 & 0.316 & -0.239 & 315.113 \\
\textit{RF}           & 192 & 43.750 & 0.316 & -0.181 & 293.885 \\
\textit{buy and hold} & 388 & 45.619 & 0.344 & -0.255 & 386.000 \\

\midrule
& \multicolumn{5}{c}{ES} \\
& num. trades & winrate[\%] & $\hat{\mu}_{r>0}$[\%] & $\hat{\mu}_{r<0}$[\%] & avg. holding period [min.] \\ \hline

\textit{RIF}          & 200 & 55.500 & 0.561 & -0.486 & 383.505 \\
\textit{RF}           & 396 & 56.566 & 0.302 & -0.271 & 180.364 \\
\textit{buy and hold} & 388 & 53.213 & 0.488 & -0.441 & 386.000 \\ 

\midrule

& \multicolumn{5}{c}{CL} \\
& num. trades & winrate[\%] & $\hat{\mu}_{r>0}$[\%] & $\hat{\mu}_{r<0}$[\%] & avg. holding period [min.] \\ \hline

\textit{RIF}          & 164 & 48.171 & 1.011 & -1.038 & 381.970 \\
\textit{RF}           & 188 & 42.553 & 0.365 & -0.435 & 154.261 \\
\textit{buy and hold} & 388 & 38.196 & 0.951 & -1.041 & 386.000 \\

\bottomrule
\end{tabular}
\vspace{5pt} 
\caption{Out-of-sample trade statistics of the proposed approach (\textit{RIF}), in comparison with the reinforcement learning strategy (\textit{RL}) and the buy and hold trades, for the three considered futures contracts.}
\label{tab:results_trades}
\end{table*}

\begin{figure}
    \centering
    \includegraphics[width=0.5\textwidth]{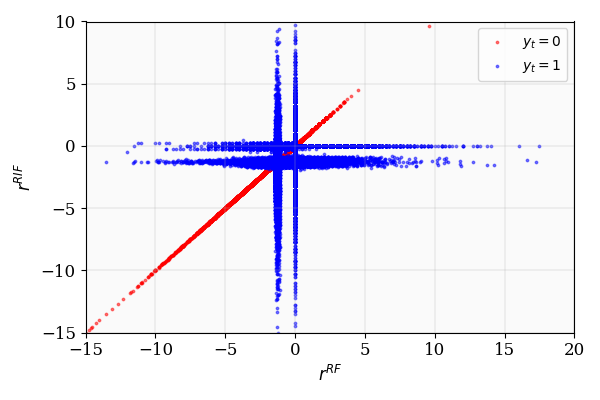}
    \caption{Proposed reward signal ($r^{RIF}$) compared to the reinforcement feedback ($r^{RF}$) over $100{,}000$ time steps of random policy evaluation on asset ES. The expert commission is set to $\vartheta = 3$ bps, while the trading commission in the experiment is set to $\phi=3$ bps.}
    \label{fig:reward_scatter}
\end{figure}

First, we compare the proposed reward signal to the reinforcement feedback. A random policy, which selects actions with equal probability, was evaluated over $100{,}000$ time steps on asset ES. At each time step, the reinforcement feedback ($r^{RF}$) and the proposed reward signal ($r^{RIF}$) were calculated. The scatter plot of $r^{RIF}$ versus $r^{RF}$ is shown in Fig.~\ref{fig:reward_scatter}.  When the oracle label is out of position ($y_t = 0$), the reward signals are the same. Otherwise, when the oracle label is $1$, several patterns emerge.
Firstly, when $r^{RF}$ is zero or negative in the magnitude of commission costs, $r^{RIF}$ reflects negative values when the agent fails to capitalize on uptrends, or positive values when it successfully avoids downtrends (vertical lines).
Conversely, when $r^{RF}$ is different from zero, $r^{RIF}$ is zero or negative in the magnitude of commission costs as the agent aligns with the oracle labels (horizontal lines). 
The plot demonstrates a mapping from reinforcement feedback to the new reward signal space, effectively providing a more stable reward structure.

The proposed approach is tested as described in the previous section and compared against two benchmarks: (1) an RL agent trained solely with reinforcement feedback and (2) a passive benchmark strategy (buy and hold). We also benchmarked our approach against an agent trained solely with imitation feedback, analogous to the standard IL method of BC. However, due to the inherent limitations of BC within the supervised learning paradigm -- specifically, the issue of distributional shift \cite{Ross2011} and the absence of an exploration-exploitation trade-off since the reward signal does not depend on the agent's actions -- the BC agent accumulated significant generalization errors. Consequently, the agent failed to achieve acceptable financial performance metrics. As these results were highly unsatisfactory and not comparable to the other benchmarks, they were omitted from the tables presented. The buy and hold benchmark involves entering a long position at the beginning of the trading day and exiting at the end, serving as a simple yet effective and most popular benchmark for performance assessment. These benchmarks are particularly suitable for our experimentation because they provide a clear contrast between different trading approaches. By comparing our agent to the RL agent with only reinforcement feedback, we can isolate the impact of the imitation feedback from the oracle labels. This comparison helps us understand whether the integration of expert demonstrations in the reward function leads to more effective policy learning. The buy and hold strategy allows us to gauge the practical profitability of our approach in real-world noisy trading conditions. If our agent can outperform this passive strategy, it indicates that the agent is not only learning from the expert demonstrations but also exploiting short-term market opportunities to achieve higher returns.

We first inspect the statistics of the trades made by the different approaches, reporting the number of trades, winrate percentage, the average return of the trades ending in profit (average positive return) and the average return of the trades resulting in losses (average negative return). We also report the average holding period (in minutes) over all trades. Table \ref{tab:results_trades} shows these statistics for all three assets. The total number of trading days in the test period (from October 2022 to March 2024) for all three assets was 388. The number of trades made by the proposed approach combining reinforcement and imitation feedback (RIF) is drastically reduced for ES, modestly for CL, and slightly increased for GC, when compared to the reinforcement feedback approach (RF). The number of trades of the buy and hold strategy is always equal to the number of days in the test period since it only holds the assets during the market open time. Although the number of trades is not uniformly reduced across assets, the average holding period of the trades is generally always longer than RF, affirming the more stable nature of the decisions made by the proposed approach. Moreover, the winrate is generally increased when compared to the buy and hold trades and is improved over the RF winrate in 2 out of 3 test assets. Finally, even though the unprofitable trades are not necessarily less risky in the RIF approach, the profitable trades themselves are shown to be more profitable on average, as suggested by the mean positive returns.

Since the strategies operate on an intraday basis, we calculate the daily returns and annualize the profitability and risk-adjusted return measures accordingly. In our results, we report the annualized mean return $\hat{\mu}_r$ and volatility $\hat{\sigma}_r$ of the strategy returns, together with the maximum drawdown (mdd) and the Sharpe ratio $\hat{\mu}_r/\hat{\sigma}_r$ (assuming the risk-free rate is zero). Table \ref{tab:results_returns} shows the out-of-sample return statistics for the strategies obtained using the proposed RIF approach, compared to the strategy built using only the RF approach and the buy and hold returns. It is evident that the RIF approach universally improves over both benchmarks in terms of returns. However, in certain assets, the approach seems to take on slightly more risk, as measured by the volatilities and maximum drawdowns. Nevertheless, this risk seems to be well compensated by the increased return, as suggested by the higher Sharpe ratios in all tested cases.

\begin{table}[]
\centering
\begin{tabular}{r|rrrr}
\toprule
& \multicolumn{4}{c}{GC} \\
& $\hat{\mu}_r$[\%] & $\hat{\sigma}_r$[\%] & mdd[\%] & Sharpe \\ \hline
\textit{RIF}          & 3.388 & 4.897 & 3.881  & 0.692 \\
\textit{RF}           & 2.012 & 4.264 & 4.606  & 0.472 \\
\textit{buy and hold} & 1.319 & 6.551 & 11.246 & 0.201 \\               
\midrule

 & \multicolumn{4}{c}{ES} \\
& $\hat{\mu}_r$[\%] & $\hat{\sigma}_r$[\%] & mdd[\%] & Sharpe \\ \hline
\textit{RIF}      & 10.982 & 8.609  & 6.554 & 1.276 \\
\textit{RF}       & 8.823  & 7.361  & 4.559 & 1.199 \\
\textit{buy and hold} & 10.348 & 10.433 & 7.237 & 0.992 \\

\midrule

 & \multicolumn{4}{c}{CL} \\
& $\hat{\mu}_r$[\%] & $\hat{\sigma}_r$[\%] & mdd[\%] & Sharpe \\ \hline
\textit{RIF}      & -6.528  & 13.710 & 27.016 & -0.476 \\
\textit{RF}       & -12.767 & 7.452  & 22.073 & -1.713 \\
\textit{buy and hold} & -22.905 & 21.088 & 39.901 & -1.086 \\

\bottomrule
\end{tabular}
\vspace{5pt} 
\caption{Out-of-sample return statistics of the proposed approach (\textit{RIF}), in comparison with the reinforcement learning strategy (\textit{RL}) and the buy and hold returns, for the three considered futures contracts.}
\label{tab:results_returns}
\end{table}

These results affirm the proposed approach and suggest that the combination of the reinforcement and imitation feedback in the reward function does indeed improve the performance of the agent. The decisions made (trades) are more stable, yielding longer holding periods which manage to capture the trends which may take more time than given by the reinforcement feedback approach. The number of trades is either commensurate or drastically reduced (in the ES case), while the profitability of profitable trades is increased. This results in higher average returns and strategies that are better compensated for the risk taken.

\section{Conclusion}\label{Conc}

The relatively low signal-to-noise ratio in financial markets leads to noisy estimates of the reward function, which hinders effective policy learning by reinforcement learning (RL) agents. To address this problem, we present a novel approach to improve RL in financial asset trading by incorporating imitation learning feedback into the reward function. By using a trend labeling algorithm as an expert, the proposed reward signal combines both reinforcement (agent's) and imitation (expert's) feedback to create a more robust reward function. 

The experimental results showed that the proposed method significantly outperforms traditional RL agents trained solely with reinforcement feedback and the passive buy and hold strategy. The inclusion of imitation feedback led to more stable trading decisions, longer holding periods, and an overall reduction in the number of trades. Moreover, the proposed approach achieved higher win rates and a higher mean value of positive returns relative to negative returns. In terms of profitability and risk-adjusted return measures, the proposed method consistently delivered higher returns and Sharpe ratios compared to the benchmarks. Although some increase in risk was observed for certain assets, this risk was well compensated by the enhanced returns. These results confirm the value of incorporating imitation feedback into the reward function and show improved performance under noisy trading conditions.

Overall, this work paves the way for more robust model-free RL applications in financial decision-making. Future research could extend these results by applying the proposed framework to a wider range of assets and market conditions, further optimizing the imitation feedback mechanism, and exploring the integration of other data sources to increase the robustness of trading strategies.

\section*{Acknowledgment}
This research has been supported by the European Regional Development Fund under the grant PK.1.1.02.0008 (DATACROSS).

\bibliographystyle{IEEEtran}
\bibliography{sample-bibliography} 

\end{document}